# A Survey on non-English Question Answering Dataset

Andreas Chandra*, Affandy Fahrizain*, Ibrahim*, Simon Willyanto Laufried*

Jakarta Artificial Intelligence Research

**Abstract**

Research in question answering datasets and models has gained a lot of attention in the research community. Many of them release their own question answering datasets as well as the models. There is tremendous progress that we have seen in this area of research. The aim of this survey is to recognize, summarize and analyze the existing datasets that have been released by many researchers, especially in non-English datasets as well as resources such as research code, and evaluation metrics. In this paper, we review question answering datasets that are available in common languages other than English such as French, German, Japanese, Chinese, Arabic, Russian, as well as the multilingual and cross-lingual question-answering datasets.

## 1 Introduction

Question answering system is a task where the model is given a question and predicts the correct answer from large data sources such as Wikipedia, books, documents, and the web. Question answering system is one of the most challenging tasks in NLP and has received massive attention from researchers and practitioners over the past decades. One of the approaches to tackle this problem is using machine reading comprehension, and the other is an open-domain question answering.

With the recent advances in deep learning methods especially for natural language processing and understanding as well as the amount of data that can be used for any NLP tasks, the question-answering research field is also experiencing a huge amount of research progress such as publicly accessible datasets, competitions, leaderboards, and more challenging tasks. This trend leads to several survey papers that capture and analyze existing datasets that are available and map the progress in QA research. However, a comprehensive review of existing resources is insufficient in the non-English datasets.

There are some questions answering and machine reading comprehension survey papers that really focus on English datasets and approaches and a brief overview of multilingual and cross-lingual resources. Previous studies have conducted the survey in the area of machine reading comprehension (Liu et al., 2019b; Zeng et al., 2020a; Dzendzik et al., 2021) give a comprehensive overview of machine reading comprehension datasets, statistics and further research directions as well as an in-depth analysis of the existing architectures to solve the task. Those surveys also give a definition of machine reading comprehension tasks and the issues that remain exist. Some previous papers also pay attention to the main question-answering systems such as (Cambazoglu et al., 2020; Huang et al., 2020; Rogers et al., 2021) give a thorough review of English question-answering dataset as well as the methods and architectures. These surveys are well-known for researchers as a pointer to conduct further research in this area.

Different from previous survey papers, this study provides a comprehensive survey on machine reading comprehension and question-answering in the non-English datasets. To the best of our knowledge, none of the existing literature examines this problem. We further analyze how the dataset is made, the type of tasks, characteristics, and challenges as well as provide a comprehensive

---

* equal contribution correspondence to andreas[at]jakartaresearch[dot]com

table of metrics in each dataset. Finally, we provide a short discussion of the existing dataset and recommendations for further development.

## 2 Data Collection

In this section, we describe some common approaches in order to build a dataset in the non-English dataset. (Rajpurkar et al., 2016) created an influential work on how they build SQuAD dataset which utilizes Wikipedia and human annotator to generate question-answer pairs. There are mainly three approaches to build machine reading comprehension or question answering datasets, original source with human-generated, professional translation, and machine translation which each approach has its own advantages, disadvantages, and challenges.

**Original source**, SQuAD dataset has the biggest influence on other researchers to build their own dataset in their own language, we found many papers that follow this style with the same source as SQuAD using Wikipedia such as in German (Cramer et al., 2006; Risch and Pietsch, 2017) French (d'Hoffschmidt et al., 2020; Heinrich et al., 2021; Keraron et al., 2020) Arabic (Mozannar et al., 2019), Hebrew (Keren and Levy, 2021), Chinese (Cui et al., 2020; Shao et al., 2018), Vietnamese (Nguyen et al., 2020a), and Japanese (Takahashi et al., 2019), and Multilingual dataset from Wikipedia source (Clark et al., 2020; Liu et al., 2019a). Another source of dataset is from the school examination which is created for the specific grade and provided passage, question, and answer. This type of data source is usually for multiple-choice tasks and some researchers made MRC datasets this way (Hardalov et al., 2019; Sayama et al., 2019; Anuranjana et al., 2019; Nguyen et al., 2020b). Medicine documents (Zhang et al., 2021b; van Nguyen et al., 2020; Veisi and Shandi, 2020). From knowledge bases (Korablinov and Braslavski, 2020; Rybin et al., 2021; Etezadi and Shamsfard, 2020). Using a search engine to construct a dataset (Ismail and Homsi, 2018; He et al., 2018). And others from book and news websites (Cui et al., 2016; Cui et al., 2019).

**Machine translation**, because acquiring data source, annotating, and building a high-quality dataset is costly, some researchers translate existing datasets like SQuAD into their own language. (Croce et al., 2018; Atef et al., 2020; Abadani et al., 2021; Lee et al., 2019). But this will cause another problem, the dataset that is from native English is English centric such as American cultures, politics, events, the history which does not reflect information in other languages. Another is lost in translation that is not able to translate properly.

**Human translation**, to avoid those problems made by machine translation, this type of dataset construction is mainly used for multilingual and cross-lingual datasets which tackle a more challenging task. This is required in order to train the model to be able to answer the question in multiple languages. Moreover, some studies construct a dataset from native speakers which actually represents their culture and interest as in (Siblini et al., 2021; Longpre et al., 2020; Asai et al., 2021).

## 3 Tasks

These are four tasks that are mainly formed in the machine reading and question-answering dataset. Those tasks are multiple-choice, cloze style, span-based, and free form. Each task will have its own challenges and approach to tackle this task. The multiple-choice form is the simplest form of question answering while the free form is the most difficult task.

### 3.1 Multiple Choice

(Biology) The thick coat of mammals in winter is an example of:

A. physiological adaptation
B. behavioral adaptation
C. genetic adaptation
D. morphological adaptation

Figure 1: Example of a multiple-choice question and answer. The correct answer is highlighted.

In the multiple-choice task, the datasets consist of questions Q and for every question, there are paired with a set of options O = {O1, O2, ... On}. Therefore, the model has to predict the correct answers given the options. Many multiple-choice datasets are derived from real-world examinations, such as the Bulgarian Reading Comprehension Dataset (Hardalov et al., 2019), ParsiNLU (Khashabi et al., 2021), and ARC (Clark et al., 2018). Figure 1 is an example of a multiple-choice task dataset taken from the Bulgarian Reading Comprehension Dataset.

## 3.2 Cloze task

**Context**
The BBC producer allegedly struck by Jeremy Clarkson will not press charges against the "Top Gear" host, his lawyer said Friday. Clarkson, who hosted one of the most-watched television shows in the world, was dropped by the BBC Wednesday after an internal investigation by the British broadcaster found he had subjected producer **Oisin Tymon** "to an unprovoked physical and verbal attack." . . .

**Query**
Producer **X** will not press charges against Jeremy Clarkson, his lawyer says.

**Answer**
Oisin Tymon

Figure 2: Example of a cloze-style dataset from (Hermann et.al., 2015)

Cloze style is a common way to examine a person's language proficiency. Usually, people who take this test will be provided with a passage that contains missing words that have to be filled. In the cloze-style question answering tasks, a dataset usually consists of paragraphs that contain missing words, the candidate's words for each blank spot, and answers derived from the candidate words.

Datasets for the cloze task can be built using two methods: automatic and manual generation processes. In the automatic process, the missing words are placed in the paragraphs randomly or using some predefined rules. For example, in Children's Book Test (Hill et al., 2016), each question contains 21 sentences where the first 20s are the context and the last sentence contains a missing word which must be filled using the context and the candidate answers provided. In the manual process, for instance, in the CLOTH dataset (Xie et al., 2018), the author utilized real English proficiency tests for Chinese students in high schools which were designed by English teachers. These exams are then converted to texts using OCR methods. The challenge in building manual process datasets is they still require human intervention such as removing invalid questions or cleaning the results from the OCR process.

## 3.3 Span-based

The development of plate tectonics provided a physical basis for many observations of the solid Earth. Long linear regions of geologic features could be explained as plate boundaries. Mid-ocean ridges, high regions on the seafloor where hydrothermal vents and volcanoes exist, were explained as **divergent boundaries**, where two plates move apart. Arcs of volcanoes and earthquakes were explained as **convergent boundaries**, where one plate subducts under another. Transform boundaries, such as the San Andreas fault system, resulted in widespread powerful earthquakes. Plate tectonics also provided a mechanism for **Alfred Wegener**'s theory of continental drift, in which the continents move across the surface of the Earth over geologic time. They also provided a driving force for crustal deformation, and a new setting for the observations of structural geology. The power of the theory of plate tectonics lies in its ability to combine all of these observations into a single theory of how the lithosphere moves over the convecting mantle.

What is the area called where two plates move apart?
divergent boundaries

What is the area called where one plate subducts under another?
convergent boundaries

Whose theory was the theory of continental drift?
Alfred Wegener

Figure 3: Example of span-based dataset taken from SQuAD2.0. The colored text is the answer to the following question.

In reading comprehension, we are often asked to select words or sentences in the paragraph as the answer to a question. This kind of answer is called span. In span-based QA, the machine needs to predict the position of the beginning and the end of the span in the given context (Zeng et al., 2020b).

The dataset creation process involves crowd workers making questions based on the particular paragraph and highlighting a span in the paragraph considered as an answer. Some examples of span-based dataset are (Rajpurkar et al., 2016; d'Hoffschmidt et al., 2020; Nguyen et al., 2020a; Lim et al., 2019). Figure 3 shows the example of a span-based dataset taken from SQuAD 1.1.

### 3.4 Free form

…Jan - I have some very upsetting news.
Last night one of our students, Malcolm Kaiser
took his own life, for those of you who knew him, there will be a memorial service at
Dempsey hill on Friday. I know it hurts, it's painful to lose someone...

Who commits suicide?
Malcolm

Figure 4: Example of free-form answer taken from NarativeQA.

In a free-form answer task, the machine can generate any answer even if the words or sentences do not exist in the paragraph. Free form can mean anything like words, sentences, images and it can come from anywhere (Zeng et al., 2020b). The examples of free-form answer datasets are (He et al., 2018; Sumikawa et al., 2019).

## 4 Datasets

In this section, we will discuss the existence of machine reading comprehension as well as question-answering datasets across languages for the non-English datasets. We reviewed some of the main languages in the world such as Chinese, Japanese, Vietnamese, Korean, Hindi, France, Russian, German, Italian, Bulgarian, Portuguese, Arabic, Persian, Hebrew.

### 4.1 Chinese

Mandarin Chinese is the second largest language with about 1.1 billion speakers all over the world. Mandarin MRC-QA dataset is considered a high resource language dataset in the non-English dataset as the dataset is abundant and consists of many types. (Cui et al., 2016) People Daily is an MRC-QA dataset on modern Chinese. It is a Cloze-style MRC that takes data from a newspaper, People Daily. The datasets are being automatically annotated with each answer having to be a noun and appearing at least twice in the passage. As an out of sample data another dataset called Children's fairy tale (which covers an entirely different topic from the news) is used and processed just like People daily. A method also being proposed called CAS reader which uses consensus to predict the answer. The result for people's daily dataset is 68.1 % accuracy, and for the Children's fairy tale dataset yields 41.3% accuracy. In 2017, (He et al., 2018) proposed DuReader. It is a QA-MRC dataset in modern Chinese, but the task type is Free form answers from the Baidu search engine. Answer annotations are done using crowdsources, where the workers are given the questions and a set of relevant documents and they must summarize the answer with empty answer options, and multiple answers for 1 question. Multiple answer questions make up about 70% of the question and about 10% with no answer mark. In total, the datasets consist of 1 million articles with 200 thousand questions and 420 thousand answers. The modeling of the QA is using Match LSTM where the answer is a span of sentences in the paragraph with BLEU-4% this algorithm resulted in 31.8% and 39.2% with Rouge-L% as comparison human performances are 56.1% in BLEU-4% and 57.4% in Rouge-L%.

CMRC (Chinese Machine Reading Comprehension) is an evaluation workshop held by China National Conferences on Computational Linguistics. During the workshop, each team will compete to make the best models on a dataset. Currently, there are 2 Datasets being published, there are from CMRC in 2017 and CMRC in 2018. In CMRC-2017 (Cui et al., 2019) there is a QA MRC dataset in modern Chinese. It is a close-style MRC datasets and a span-based datasets. The close-style datasets are automatically annotated using a computer program with only nouns and entity answers accepted, for the span-based datasets are manually annotated with 5 question per passage. In total there are 20000 articles for the cloze-style datasets and about 5000 paragraphs for the user-query datasets. These datasets are used for the competition and the winner is 6ESTATES team for cloze-style datasets with 81.9% accuracy beating the baseline of AoA reader with 78.63% accuracy. The user-query datasets winner is ECNU with 69.53% accuracy beating the baseline of AoA reader with 51.53% accuracy. In (Cui et al., 2020)

CMRC-2018 it is a QA-MRC Dataset using modern Chinese. It is a span-based answer MRC which takes data from Wikipedia. Each Wikipedia article is divided into a passage and each passage is annotated by experts. Experts are required to generate a maximum of 5 questions per passage and each answer has no more than 30 Chinese characters. The datasets are divided into train, test, dev, and challenge (a set with higher difficulty). For the test, dev and challenge set each question required 3 answers. These datasets consist of around 20,000 Questions without further information about passages and answers. To test the datasets, a neural network algorithm called Z reader performs the best with EM accuracy around 74.175% and F1 accuracy around 88.145%, with baseline of human performance on EM accuracy around 92.4% and F1 accuracy around 97.914%.

Historically, the Chinese Language is divided into simplified Chinese and Traditional Chinese. Traditional Chinese has a more complex expression on each character and often appears in old texts or stories. There is one dataset that we observed offers a QA-MRC dataset in this traditional Chinese called DRDC (Shao et al., 2018). It is a span-based answer MRC which takes data from Wikipedia. Question generation is crowdsourced with each worker required to generate 3-5 questions per passage along with its answer. Dominantly question type with "which" are the most on this dataset, and answer in type of "entity". All these datasets consist of 2,108 articles, 10,014 paragraphs, 200,000 questions and 400,000 answers. The model generation technique is using a transfer learning BERT and scored 82.34% and 89.59% for EM and F1 respectively. In addition, Human performance is higher in F1 compared to the model that gets 80.43% and 93.3% for EM and F1 respectively.

Other than General QA we also included a medical domain QA-MRC Dataset called CMedRC (Zhang et al., 2021b). In the Medical domain QA jargons and loanwords are often used which implies it is a harder type of dataset. CMedRC is a span-based answer datasets from a website called DXY Medical China. Question answer generation is using a group of medical experts where each expert will cross check their peer's pair each. This dataset consists of 18153 medical related questions and modeled on several neural network and transfer learning algorithms. CMedBERT which the proposed resulted in 72.91% EM accuracy and 85.64% F1 accuracy. As a comparison Human perform in 85% EM and 96.69% F1 There are also some analyses which explain how the machine fails on some of the questions which is Long-tail Terminology where there are 2 terminologies used in medical terms and some terminology combinations where 2 words give different meaning when combined. Another domain specific QA-MRC is on sport domain, Live QA (Liu et al., 2020). This paper proposes a question-answering dataset constructed from the play-by-play live broadcast. The dataset contains 117K multiple-choice questions written by human commentators for over 1,670 NBA games. The dataset provides a new challenge in the area of question-answering research by focusing on the ability to reason across timeline-based live broadcasts.

### 4.2 Japanese

Japanese is a character-based language like Chinese and Korean. The Kanji (Japanese alphabet) share and borrow many characters from traditional Chinese. The Japanese QA-MRC dataset is not as abundant as Chinese, but we found a couple of interesting datasets. (Nishida et al., 2018) JP-News is a QA MRC dataset using Japanese. It is a span-based answer dataset that is crowdsourced from Japanese Newspapers. Each worker is required to make 5 questions per article. These datasets are tested using a method proposed by the author which combines the process of Information retrieval and reading comprehension. Since it is a news article there is a chance on the IR process several duplicate articles were selected. In total, the datasets consist of 14804 articles with only 5000 samples and transform to be 12485 paragraphs with 82592 questions and 224908 answers. For performances in terms of Information Retrieval (IR), datasets achieve 83% accuracy using MTL and for RC EM accuracy reaches 78.8% and 89.2% F1 accuracy using the same MTL algorithm.

Predicate Argument Structure is a structure of syntactic and semantic relations between words in sentences. (Takahashi et al., 2019) PAS-QA is the only dataset that we found focusing on PAS in the form of QA dataset. To enhance performances of PAS-QA, this paper also made a Reading Comprehension dataset RC-QA. PAS-QA consists of 2146 articles from the driving experience corpus which transforms into 17290 questions. On the other hand, RC-QA consists of 5146 articles and 20007 questions from the same Corpus. Question answer generation is crowdsourced and cross-checked. This combination of models resulted in 78% accuracy on nominative structure, 53% on the accusative structure, and 51% on the dative structure using the MC-stepwise algorithm.

(Sumikawa et al., 2019) proposed a dataset called e-learning FAQ which aims to solve Chabot in MRC fashion which is used specifically in an e-learning system. It is a free-form answer dataset with questions that are real questions from e-learning users of Tokyo Metropolitan University and the answers are provided by the e-learning system engineer. During the development of data, (Sumikawa et al., 2019) successfully collects 200 question-answer pairs, and for every semantically similar answer, the answer is required to have five questions, otherwise the question would be paraphrased. This results in total of 427 questions and 79 unique answers.

### 4.3 Vietnamese

In Vietnamese, there are three literatures that are found in developing machine reading comprehension datasets. (Nguyen et al., 2020b) proposed ViMMRC which consists of multiple-choice dataset from school supplements of grade 1 to 5. Question and answer are generated by a teacher with at least 5 questions per article with each question having 3 wrong answers and 1 real answer. In the dataset about 36.73% questions require multiple sentences reasoning. In total there are 417 articles and 2783 questions. Lexical-based approaches is used for developing a model which essentially sliding windows and distance score. Some similar semantic words for each token are also added from Wikipedia to enhance the result. Overall, the model gives 61.81% accuracy compared to 91% accuracy of Human performances.

On the other hand, the (Nguyen et al., 2020a) UIT-ViQuAD is a span-based answer QA MRC dataset which uses Wikipedia as a data source. Similarly from SQuAD, the annotation utilizes crowdsourcing that each worker is required to generate a question and selecting a span where the answer is in the passages. Overall "What-type" questions are making up most of the question data and "inference-type" answers for 68.92% of the answers, thus (Sumikawa et al., 2019) claimed these datasets are harder compared to SQuAD. The dataset contains 174 Articles, 5,109 passages, and 23,000 question-answer pairs. Several Neural Network and Transfer Learning method are experimented to examine the dataset. XLM-Rlarge gives the best accuracy (EM=68.98% and F1=87.02%) and the human performance give EM and F1 a result of 85.59% and 94.69% respectively.

From the health domain there is ViNewsQA-2020 (van Nguyen et al., 2020), The dataset is a span-based answer dataset from VnExpress health section articles. Question answers are generated by crowd workers. With most types of questions is "what" and most types of answer is logical paraphrasing. Overall, the data consist of 4,416 articles and 22,057 questions. Performances are calculated using several neural networks and transfer learning algorithms. ALBERT performs the best on the dataset which yields an EM of 65.26% and 84.89% on F1, although it still can't beat human performances on 79.79% EM and 95.79% F1.

### 4.4 Korean

The Korean alphabet, often called Hangul, is a character-based language like Mandarin and Kanji. Unfortunately, we can't find many datasets on Korean Hangul. There is one dataset called KorQuad (Lim et al., 2019), just like SQuAD it is a span-based answer dataset from Korean Wikipedia. Questions and answers are generated by 3 crowd workers. In general, the questions are divided into 6 categories, there are synonyms, general knowledge questions, multiple reasoning questions, logical questions, and bad questions. The answer is also divided into categories there are objects, people, date, location, method, and reason. Overall, there are 2980 articles, 11268 paragraphs and 70,079 questions. In performance the modeling approach is using transfer learning, using BERT it yields 71.68% EM accuracy and 89.76% F1 accuracy for comparison human performances are 80.17% in EM and 91.2% in F1.

### 4.5 Hindi

Hindi is the number 3 on the list of most spoken languages in the world with about 342 million speakers all over the world, but The MRC QA dataset is hardly found. We believe it is due to the English language being largely used as the professional language in India. HindiRC (Anuranjana et al., 2019) is a QA-MRC Dataset using Hindi (Indian language). It is a span-based answer MRC that takes data from children's learning supplements from grade 2-5 in India, consequently, the dataset will experience a gradual increase in difficulty from each grade. Question and answer pairs are labeled by experts and taken from education websites. The datasets consist of 24 passages in total with 127 pairs of questions and answers. In addition, the paper benchmark several methods for textual similarity task and BM25 outperform compare to other methods.

### 4.6 French

FQuAD (d'Hoffschmidt et al., 2020) is a French MRC dataset collected from 1,769 articles of French Wikipedia pages. This dataset comes in two versions, FQuAD1.0 and FQuAD1.1. The 1.0 version built from 145 randomly sampled articles divided into training, development, and test sets with a total 117, 18, and 10 articles respectively. Those quantities resulted in 26,108 QA pairs in total consisting of 20731 training sets, 3188 development sets, and 2189 test sets. While there are 181 additional articles for 1.1 version resulting in 326 articles being used for this version. The training, development, and test set for this version consist of 271, 30, and 25 articles. The 1.1 version has more difficult questions with 62,003 QA pairs consisting of training, development, and test sets are respectively 50741, 5668, and 5594 pairs. Both datasets followed SQuAD1.1 format and data collection process as well. The evaluation process also ignores the following articles: le, la, les, l', du, des, au, aux, un, une. Baseline Human performance score reaches 92.1% F1 and 78.4% EM on FQuAD1.0, and 91.2% F1 and 75.9% EM on FQuAD1.1.

At the same time, (Keraron et al., 2020) released French QA datasets named PIAFv1.0. This dataset is based on French Wikipedia pages having the same PageRank metrics as SQuAD. PIAFv1.0 consists of 3835 QA pairs generated from 761 paragraphs annotated by 285 PIAF team volunteers. It then performed comparative analysis between PIAFv1.0 and FQuAD using CamemBERT trained on combination dataset between SQuAD-Fr, FQuAD, sub-sample of FQuAD, and PIAFv1.0. The result shows that combining PIAFv1.0 with another dataset gives a comparable improvement. Hence, The paper claimed that PIAFv1.0 might be effectively used for training data augmentation.

While recent models trained on existing dataset cannot understand when a given question has no answer, therefore (Heinrich et al., 2021) then introduced FQuAD2.0, the first non-English native datasets for adversarial QA. FQuAD2.0 extends FQuAD with 17.765 unanswerable questions divided into training, development, and test sets in total respectively 9481, 4174, and 4110. Entity swap, negation, or something else used to generate unanswerable questions. Each annotator has to generate at least 4 adversarial questions. The study introduced two other metrics: F1has ans and NoAnsF1 which respectively are the average F1 score of answerable questions and F1 score of determining whether a question is unanswerable. CamemBERT Large being used as FQuAD2.0 baseline obtained score for EM, F1, F1hasAns, NoAnsF1 metrics are 78%, 83%, 90.1%, and 82.3% respectively.

### 4.7 Russian

The Russian dataset is one of low resource languages compared to the English dataset. As an effort to support development of QA in Russian, (Korablinov and Braslavski, 2020) released RuBQ, a Russian KBQA dataset curated from Russian quiz collections website. The data collection process discarded any QA pairs that don't contain any Russian words and phrases, also crossword questions asked the number of letters in the expected answer. This dataset contains 1500 unique QA pairs with 300 of them are unanswerable questions. DeepPavlov, the KBQA implementation for Russian, and WDAqua, the multilingual rule-based KBQA system, are being used as baseline models. The result shows WDAqua outperforms DeepPavlov in terms of precision on the answerable subset with score 16% and 13% respectively. On the other hand, it has a lower accuracy on unanswerable questions with scores 43% and 73% for WDAqua and DeepPavlov respectively.

A year after that, (Rybin et al., 2021) released RuBQ2.0 which expands the first version with Yandex and Google search logs. By applying crowdsourcing to annotate data, this extension dataset is almost twice bigger than the first version that is 2910. The unanswerable questions are also increased by 210 from 300, resulting in RuBQ2.0 has in total 510 unanswerable questions out of 2910 total data. But the dataset format is different with SQuAD since crowdworkers mark a paragraph instead of a span as an answer. DeepPavlov, QAnswer-Russian, QAnswer-English, and Simple baseline are being used for KBQA, where total corrected answer for each model is 683, 541, 570, and 841 indicating both DeepPavlov and QAnswer cannot outperform Simple baseline. Apart from the RuBQ family, there is another Russian dataset called SberQuAD (Efimov et al., 2020). This dataset is based on the Reading Comprehension task of the Sberbank 2017 competition with total 50364 paragraph-question-answer triplets. Interestingly, SberQuAD does not tell which Wikipedia pages a paragraph belongs to and each answer is only represented by a string without its respective position as we found in SQuAD. Using the same metrics as SQuAD that is EM and F1, it can be found that the BERT-based QA model by DeepPavlov reached the highest score for EM and F1 respectively 66.6 and 84.8.

### 4.8 German

CLEF series is the only German QA dataset evaluation with 200 pairs per year back then. (Cramer et al., 2006) then proposed a method to create German open-domain QA dataset by harvesting Wikipedia articles. The method specifically marks a span containing some fact, then write a question about it. By taking a snapshot on a two-month time window, the proposed method can collect 1454 QA pairs which is more than CLEF that only released 200 pairs per year. Lately, (Risch and Pietsch, 2017) released GermanQuAD dataset consisting of 13722 extractive QA pairs mined from German version of English Wikipedia articles used in SQuAD. The dataset splits into training and test sets. Training set has in total 11518 QA pairs where each question has only one answer. On the other hand, the test set has three answers per question resulting in a total 2204 questions and 6536 answers. The baseline human performance evaluation result was 66.4 and 89.5 for EM and F1 metrics.

### 4.9 Italian

(Croce et al., 2018) proposed an Italian QA dataset by automatically translating an existing English dataset called SQuAD-IT which has a larger size than the existing Italian QA dataset. The training set contains 18506 paragraphs, 54159 questions and answers pairs. While the test set contains 2010, 7609, and 21489 for paragraphs, questions, and answers respectively. DrQA-IT models were used to evaluate the created dataset. The model scored 56.1 and 65.9 for EM and F1.

### 4.10 Bulgarian

(Hardalov et al., 2019) introduced a new Bulgarian reading comprehension dataset containing 2636 multiple-choice type questions collected from matriculation exams and online quizzes. The topic covered in this dataset is biology, philosophy, geography, and history. In total 2221 questions were collected from matriculation exams, while the online quizzes resulted in a total of 412 questions. The multilingual BERT trained on the RACE dataset scored an accuracy of 42.23.

### 4.11 Portuguese

FaQuAD (Sayama et al., 2019) is a machine reading comprehension (MRC) dataset in the domain of Brazilian higher education. The dataset was collected from 18 official documents of computer science colleges in Brazil and 21 Wikipedia articles related to the Brazilian higher education system. This dataset is the first Portuguese MRC dataset which contains 900 questions. This dataset was evaluated by a BiDAF-based model provided by AllenNLP and using two different contextual representations ELMo and GLoVe. The results show using ELMo contextual representation is better than GLoVe. ELMo scored 43.9% F1 and 24.5% EM, while GLoVe scored 36.9% F1 and 17.6% EM.

### 4.12 Arabic

Arabic is one of the most spoken languages in the world with more than 300 million speakers and is adopted as an official language in more than 20 countries, mainly those which are in the Middle East and North Africa region. However, with this big community of speakers, the development of the Arabic dataset in question-answering is still behind the high resource language such as English. From our discovery, there are only three Arabic QA datasets. Most of those datasets follow the SQuAD format, hence utilize Wikipedia pages as the main source of the data.

(Ismail and Homsi, 2018) were released a dataset in 2018 called DAWQAS. This dataset covers the why QA task in Arabic. To generate the question, the author used a google search by seeding them with a "why" word, and to get the answer, they scrape the article from the search results. Before scraping the results, they made a list of credible websites to make sure the answers are legitimate. After they get the passages from the search results, the answers are selected by assigning rhetorical relations probability for each sentence in the passages. Then the answers were annotated to ensure the quality. The resulting model is 3,205 question and answers pairs.

After DAWQAS, ARCD (Mozannar et al., 2019) was released in 2019. The author closely followed SQuAD dataset-making procedure to generate the ARCD dataset. In ARCD, every tuple contains passages, questions, and answers. The passages were curated from Arabic Wikipedia pages and the questions and answers were generated by crowd workers from the Mechanical Turk platform. It has 1,395 tuples. The model performed is a BERT-based reader that scores 61.3% on the F1 score.

Like ARCD, AQAD (Atef et al., 2020) is an Arabic QA dataset that uses SQuAD as its base. This dataset is a semi-translated dataset from SQuAD, although the author claimed that it is not a translated dataset. The passages in this dataset were extracted from the respective Arabic wiki page from the English passage in the SQuAD. After they get the corresponding pages, then they filter the paragraphs in Arabic Wikipedia which matches the paragraph in English Wikipedia by using semantic meaning. After they get the paragraphs that match, the question and answer with respect to those paragraphs were directly translated from English to Arabic. AQAD contains more than 17000 tuples of passages, questions, and answers. Two models achieve 37% (BiDAF) and 32% (BERT) F1 scores.

## 4.13 Persian

The Persian language has various datasets in question-answering tasks. Although not as rich as a high resource language like English, Persian datasets for question-answering tasks utilize various techniques to generate them. Generating questions from knowledge graphs, using google search queries to generate questions, extracting questions and answers from paper-based examinations are amongst methods that are used to generate the Persian datasets.

The first QA dataset in Persian from our discovery is an unnamed dataset in a Persian medical QA system (Veisi and Shandi, 2020) which the author released in 2020. The QA system built for this dataset consisted of three main steps: questions processing, documents retrieval, and answer extraction. With this nature, this dataset has a large number of corpora that contain Persian medical articles. However, there are only 550 question and answer pairs which consist of 500 positive and 50 negative, unanswerable questions. Those question-answers pairs were posed by general users and physicians. The QA system built by the author, which combined both information retrieval and natural language processing methods has a performance of F1 score at 83.7%.

After the release of the first dataset, in the same year, ParsiNLU (Khashabi et al., 2021) was released. Compared to the first one, ParsiNLU is a more complete dataset that covers various NLU tasks. Regarding the question-answering task, ParsiNLU includes datasets for machine reading comprehension (MRC) and multiple-choice question answers (MCQA). The author generates the MRC dataset using google search queries as the questions feeder. Then, the results of the queries will be used as the contexts to answer the questions. To get the MCQA data, ParsiNLU utilizes general paper-based tests for students and employees that have multiple choice questions in there. Those exams were then scanned and converted to text using OCR techniques and annotated afterward by Persian speakers. In ParsiNLU, there are 800 examples of MRC data and 2460 examples of MCQA. The best performing model on the MRC dataset is the mT5-XL model with an F1 score of 70.4% compared to an 86.2% F1 score of human performance. For the MCQA dataset, the author divided the evaluation based on the three question topics: literature, common knowledge, and math and logic. The best performing models based on the accuracy score for each topic are WikiBERT-base (36.9%), WikiBERT-base (30.2%), and mt5-small (39.1%) respectively.

Not long after ParsiNLU, PeCoQ (Etezadi and Shamsfard, 2020) was released. The author used a different technique from the previous two where they utilize FarsBase (Asgari-Bidhendi et al., 2019), a Persian knowledge graph, to build the dataset. From all nodes and edges in the knowledge graphs which acts as an entity or relationship, the authors extracted questions and answers using some predefined syntactic rules or templates. Then, these produced pairs were curated by Persian linguists to ensure their correctness and meaningfulness. PeCoQ consisted of 10,000 question-answer pairs. However, the author did not introduce any model baseline for this dataset.

Table 1: Machine reading comprehension and question answering non-English datasets

| Dataset | Year | Language | Question | Answer | Context |
| --- | --- | --- | --- | --- | --- |
| DAWQAS | 2018 | Arabic | Google Search Queries | Article from search results | General |
| ARCD | 2019 | Arabic | Crowd workers | Crowd workers | General |
| AQAD | 2020 | Arabic | Translated from SQuAD | Translated from SQuAD | General |
| Persian Medical QA System | 2020 | Persian | Crowd workers | Crowd workers | Medical |
| ParsiNLU-MRC | 2020 | Persian | Google search queries | Articles from search results. | General |
| ParsiNLU-MCQA | 2020 | Persian | exam | exam | General, history, and math |
| PeCoQ | 2020 | Persian | Persian knowledge graph (FarsBase) | Persian knowledge graph (FarsBase) | General |
| ParSQuAD | 2021 | Persian | Translated from SQuAD | Translated from SQuAD | General |
| ParaShoot | 2021 | Hebrew | Crowd workers | Crowd workers | General |
| People's Daily | 2021 | Chinese | Automatic annotation | Automatic annotation | General |
| Children's Fairy Tale | 2016 | Chinese | Automatic and manual annotation | Automatic and manual annotation | General |
| DuReader | 2017 | Chinese | Baidu search queries | Articles from search results | General |
| CMRC-2017 | 2017 | Chinese | Crowd workers | Crowd workers | General |
| CMRC-2018 | 2018 | Chinese | Crowd workers | Crowd workers | General |
| DRCD | 2018 | Chinese | Crowd workers | Crowd workers | General |
| CMedRC | 2020 | Chinese | DXY Medical China | Human experts | Medical |
| Live-QA | 2020 | Chinese | Human commentators | Human commentators | Sport |
| JP-News | 2018 | Japanese | Crowd workers | Crowd workers | General |
| PAS-QA | 2019 | Japanese | Crowd workers | Crowd workers | Driving |
| RC-QA | 2019 | Japanese | Crowd workers | Crowd workers | Driving |
| E-Learning FAQ | 2019 | Japanese | Human experts | Human experts | Education |
| ViMMRC | 2020 | Vietnamese | Exam | Teacher Annotation | Education |
| UitViQuAD | 2020 | Vietnamese | Crowd workers | Crowd workers | General |
| ViNewsQA | 2020 | Vietnamese | Crowd workers | Crowd workers | Medical |
| KorQuAD | 2019 | Korean | Crowd workers | Crowd workers | General |
| HindiRC | 2020 | Hindi | Exam | Exam | Education |
| 27 German QA Wikipedia | 2006 | German | Crowd workers | Crowd workers | General |
| 28 SQuAD-IT | 2018 | Italian | Translated from SQuAD | Translated from SQuAD | General |
| 29 Bulgarian Multiple Choice | 2019 | Bulgarian | Exam | Exam | Bio, Geo, History |
| 30 SberSQuAD | 2019 | Russian | Crowd workers | Crowd workers | General |
| 31 FaQuAD | 2019 | Portuguese | Exam | Exam | Education |
| 32 PIAF | 2020 | French | Crowd workers | Crowd workers | General |
| 33 FQuAD | 2020 | French | Crowd workers | Crowd workers | General |
| 34 FQuAD2.0 | 2021 | French | Crowd workers | Crowd workers | General |
| 35 RuBQ | 2020 | Russian | Quiz | Quiz | General |
| 36 RuBQ2.0 | 2020 | Russian | Search logs | Crowd workers | General |
| 37 GermanQuAD | 2021 | German | Crowd workers | Crowd workers | General |

Table 2: Model performance on non-English datasets

| Dataset | Model | Exact Match | F1 |
| --- | --- | --- | --- |
| DAWQAS | - | - | - |
| ARCD | BERT | - | 61.3% |
| AQAD | BiDAF | - | 37.0% |
| Persian Medical QA | Proposed system | - | 83.7% |
| ParsiNLU-MRC | mT5-XL | - | 70.4% |
| ParsiNLU-MCQA | mT5-small | - | 39.1% |
| PeCoQ | - | - | - |
| ParSQuAD | mBERT-manual | - | 70.0% |
| ParaShoot | mBERT | - | 56.1% |
| People's Daily | CANN | 68.1% | - |
| Children's Fairy Tale | CANN | 41.3% | - |
| DuReader | Match LSTM | 31.8%(BLEU-4) | 39.2%(ROUGE-L) |
| CMRC-2017 | - | 81.9% (Cloze) | - |
| CMRC-2018 | Z-reader NN | 74.1% | 88.1% |
| DRCD | BERT | 82.3% | 89.5% |
| CMedRC | CMedBERT | 72.9% | 85.6% |
| Live-QA | NN-Gated Attention | 53.1% | - |
| JP-News | MTL | 78.8% | 89.2% |
| PAS-QA | BERT (MC-Stepwise) | 76% (Nom) 53% (Acc) 51% (Dat) | - |
| RC-QA | - | - | - |
| E-Learning FAQ | Clustering | - | 81.0% |
| ViMMRC | Lexical-based | 61.8% | - |
| UitViQuAD | XLM-Rlarge | 68.9% | 87.0% |
| ViNewsQA | ALBERT | 65.2% | 84.8% |
| KorQuAD | BERT | 71.6% | 89.7% |
| HindiRC | Similarity-based | 80% (Grade 2) | - |
| German QA Wikipedia | - | - | - |
| SQuAD-IT | DrQA-IT | 56.1% | 65.9% |
| Bulgarian Multiple Choice | Multilingual BERT | 42.2% | - |
| SberSQuAD | BERT | 66.6% | 84.8% |
| FaQuAD | BiDAF+ELMo | 24.5% | 43.9% |
| PIAF | CamemBERT | - | 71.1% |
| FQuAD | CamemBERT-large | 91.8% | 82.4% |
| FQuAD2.0 | CamemBERT-large | 78% | 83.0% |
| RuBQ | WDAqua | - | - |
| RuBQ2.0 | mBERT | - | 48.0% |
| GermanQuAD | GELECTRA-large | 68.6% | 88.1% |

Last, ParSQUAD (Abadani et al., 2021), a dataset using SQuAD as its base, was released in 2021. ParSQUAD is a Persian QA dataset that is directly translated from the English SQuAD. However, the author admits that translating SQuAD directly could lead to errors such as answer span errors or translation errors. To improve its quality, the author tried to produce two versions of datasets: one with manual corrections (24,632 examples) and the other with automatic corrections (70,560 examples). The manual dataset has an mBERT F1 score of 56%, and the automatic dataset F1 score is 70% using mBERT.

### 4.14 Hebrew

Hebrew is a language spoken in Israel and one of their official languages other than Arabic. Around the world, Hebrew is largely spoken by the Jewish community, and it's estimated only 9 million people speak Hebrew. From our discovery, there is only one Hebrew dataset for the question answering task called ParaShoot. ParaShoot (Keren and Levy, 2021) is the first Hebrew dataset on QA datasets. Its format is following SQuAD's, with passages, questions, and answers pairing. Just like SQuAD, the passages in ParaShoot are curated from Hebrew Wikipedia pages, and questions were generated by Hebrew native speakers. ParaShoot has 3038 annotated examples. The best model performance in this dataset is mBERT with a 56.1% F1 Score.

Table 3: Multilingual and Cross-lingual datasets and its question-answer generation as well as its context

| Dataset | Year | Question | Answer | Context |
|---------|------|----------|--------|---------|
| BiPaR | 2019 | Crowd workers | Crowd workers | Novel |
| MKQA | 2021 | Professional Translator | Professional Translator | General |
| MLQA | 2020 | Professional Translator | Professional Translator | General |
| MMQA | 2018 | Crowd workers | Crowd workers | Social & Sciences |
| TyDi | 2020 | Crowd workers | Crowd workers | General |
| XOR-TYDI-QA | 2021 | Human Translator | Human Translator | General |

## 5    Multilingual and Cross-lingual

Multilingual and Cross-lingual is a subfield in many natural language processing and understanding tasks. The main objective is to build a model that is able to learn multiple languages at once and be able to give a query and find the answer in another language. While deep learning models require an enormous amount of data, there are several efforts to narrow the gap. Many current studies attempt to fill this gap by creating a multilingual and cross-lingual dataset in several languages and mainly derived from the English dataset. The latest performance of transfer learning and zero/few-shot methods also make this gap even closer. We review several high influential multilingual and cross-lingual question answering datasets.

(Jing et al., 2020) published a dataset called BiPaR, Bilingual Parallel Novel-style Machine Reading Comprehension in 2019. The main contribution of the paper is that the dataset builds the triples, paragraphs, question-answer pairs in parallel in English and Chinese which makes monolingual, multilingual, and cross-lingual tasks possible. The paragraphs are constructed from the novel which requires reading comprehension skills such as coreference resolution, multi-sentence reasoning, and understanding of implicit causality. Moreover, the data consists of 3,677 bilingual paragraphs and 14,668 bilingual question-answer pairs which are generated from crowdsourcing.

(Longpre et al., 2020) published a dataset in 2021, the dataset is called MKQA which samples 10K queries from the Natural Questions dataset (Kwiatkowski et al., 2019) from English and humans translate them into 25 other languages. The main objective of this dataset is to make a fair comparison between languages, especially for low-resource languages. The data contains 10K in Arabic, Danish, German, Spanish, Finnish, French, Hebrew, Italian, Japanese, Khmer, Korean, Malay, Dutch, Norwegian, Polish, Portuguese, Russian, Swedish, Thai, Turkish, Vietnamese, Chinese (Simplified, Hong Kong, Traditional) which in total 260K question-answer pair.

(Lewis et al., 2020) published a dataset in 2020. MLQA contains QA instances in 7 languages, English, Arabic, German, Spanish, Hindi, Vietnamese, and simplified Chinese. The data source is from Wikipedia and filtered the passage that has similar meaning with other languages. The questions are generated through crowdsourcing and then translated into other languages using professional translators. The data has over 12K instances in English and 5K in each other language, with each instance, the parallel between 4 languages on average.

(Gupta et al., 2018) published a dataset in 2018 namely MMQA. This paper curates 500 articles in six different domains including Tourism, History, Diseases, Geography, Economics, and Environment from the web and mainly focuses on factoid and short descriptive questions. These articles form a comparable corpus of 250 English documents and 250 Hindi documents from the web. Question-answer is constructed into 5,495 pairs and both being in English and Hindi.

(Clark et al., 2020) publish a dataset called TyDi QA in 2020. The data contains 11 languages including English, Arabic, Bengali, Finnish, Indonesian, Japanese, Kiswahili, Korean, Russian, Telugu, and Thai. The data source is from Wikipedia and question-answers are generated naturally from native speakers. The passages are collected from the top ranked result given the question using google search. The number of pairs for train, dev, and test are 166,916, 18,670, 18,751 respectively.

(Asai et al., 2021) publish a dataset in 2021 called XOR-TYDI-QA. this paper extends open-retrieval question answering to a cross-lingual setting enabling questions from one language to be answered via answer content from another language. The paper proposes a new task called cross-lingual open-retrieval question-answering and datasets called XOR-TYDI-QA which are derived from TYD-QA. (Asai et al., 2021) created a large-scale dataset that consists of 40K information-seeking questions across 7 diverse non-English languages.

(Liu et al., 2019a) publish a dataset in 2019 called XQA. The dataset contains nine languages and English as the primary language in the training set. On the other hand, the development and test set contain eight languages which are German, Portuguese, Polish, Chinese, Russian, Ukrainian, and Tamil. The source of the data is from Wikipedia which utilized a feature called "Did you know" and collected the answer from Wiki data as the golden answers. The passages are constructed using a BM25 retriever to find relevant documents. This results in 56,279 English passage-question-answer sets. For the development and test sets, the data was successfully constructed to make a total amount of 17,358 and 16,793 question-answer pairs respectively.

## 6 Discussion

In the previous section, we have analyzed the existing literature that specifically builds a dataset for machine-reading and question answering tasks. Based on the investigation, there are some unsolved issues that remain in the recent literature.

### 6.1 Limited Contextualization

We observe that most of the current datasets are relying so much on Wikipedia which is the text that is more general in certain topics. We argue that in a real-world application, machine reading comprehension must have a general and deep understanding of the topic which may lead to the level of difficulties of questions and the passage.

We also found that most of the non-English data source is from school examinations and has a very small number of passages, questions, and answers. While in the real world, we are not expected to help students with machine reading comprehension systems. Further research should consider large-scale and diverse topics in non-English datasets and more complex topics and mimic real-world applications.

### 6.2 Limited Form of Passage

In comparison to the English dataset, we often found that the source and the form of passage are more diverse. Most of the non-English datasets rely on Wikipedia, textbooks, or other formal forms of text. We argue that this form is only applicable when the source is less noisy while most of the information found on the web is colloquial such as in forum, social media, and blogs.

### 6.3 Complex Challenges

Another area of challenge in English but not found in non-English datasets is that disfluencies. In English, there is a recent paper that studied disfluencies called DISFL-QA (Gupta et al., 2021) that emphasize contextual disfluencies in the passage and question. We suggest that the non-English dataset follow this step and add more complexities in the context and questions.

### 6.4 Low Resources

Low Resources are defined by many dimensions such as the availability of the task-specific label, the availability of unlabeled language, and the availability of auxiliary data (Hedderich et al., 2021). In this survey, since most of the papers are using labeled data, low-resources languages are defined as how much labeled data are available. Compared to the English datasets, almost all languages here are considered low-resource with inadequate labels. Label generation is usually time-consuming and costly work; thus, this low-resource problem will obstruct the progress of research on those languages. Many techniques have been proposed to generate more labeled data such as data augmentation, direct translations, etc. But this technique often comes with its own disadvantages (Hedderich et al., 2021) and varies across languages and tasks.

### 6.5 Lack Human Performances Benchmark

Human performances are defined as the evaluation scores when humans are offered to do the same task with the model using the same dataset. The human performance metric is important to understand how big the gap between human and deep learning model is. For example, in the SQuAD dataset, humans achieved an F1 score of 86.1% compared to the baseline model with a 51% F1 score (Rajpurkar et al., 2016). Using this benchmark, we could understand that there is plenty of room for improvements from the basic models. However, from the articles we have surveyed, only 10 out of 35 papers report human performance on their dataset.

### 6.6 Single-hop Reasoning

The existing non-English datasets only support single passage reasoning to answer the question. While in the real-life problem we often found questions that require information from several passages or documents to answer. On the other hand, the English dataset has HotpotQA (Groeneveld et al., 2020) which supports multi-hop reasoning. It also provides supporting facts that would make models more explainable and learn the underlying process in answering questions. Addressing this issue will advance the development of non-English datasets.

### 6.7 Numerical Reasoning

One of the other advances in the development of the English dataset is numerical reasoning which is also common in the real problem. We often found some math questions which sometimes cannot be answered directly by looking for spans in the passages, but also need some calculations to answer. MathQA (Amini et al., 2019) is an example of an English dataset that is specifically intended for solving math problems. Another example is NOAHQA (Zhang et al., 2021a) which also gives interpretable reasoning in answering questions. However, there are a few of non-English datasets addressing this issue yet.

## 7 Conclusion

This paper conducts a comprehensive survey on the recent progress of non-English machine-reading and open-domain question answering research datasets. We give the lists, analysis and table progress in depth of recent work and provide background information, challenges, and results of each dataset. This paper provides pointers for researchers and practitioners to continue the research in machine reading and question answering domain and tackle the remaining challenges in this field especially for other languages other than English. Further work should therefore include visual question answering dataset and the challenges that belong to the task.

## References

Negin Abadani, Jamshid Mozafari, Afsaneh Fatemi, Mohammd Ali Nematbakhsh, and Arefeh Kazemi. 2021. ParSQuAD: Machine Translated SQuAD dataset for Persian Question Answering. In *2021 7th International Conference on Web Research (ICWR)*, pages 163–168. IEEE, May.

Aida Amini, Saadia Gabriel, Shanchuan Lin, Rik Koncel-Kedziorski, Yejin Choi, and Hannaneh Hajishirzi. 2019. MathQA : Towards Interpretable Math Word Problem Solving with Operation-Based Formalisms. In *Proceedings of the 2019 Conference of the North*, pages 2357–2367, Stroudsburg, PA, USA. Association for Computational Linguistics.

Kaveri Anuranjana, Vijjini Anvesh Rao, and Radhika Mamidi. 2019. HindiRC: A Dataset for Reading Comprehension in Hindi. In *20th International Conference on Computational Linguistics and Intelligent Text Processing*.

Akari Asai, Jungo Kasai, Jonathan Clark, Kenton Lee, Eunsol Choi, and Hannaneh Hajishirzi. 2021. XOR QA: Cross-lingual Open-Retrieval Question Answering. In *Proceedings of the 2021 Conference of the North American Chapter of the Association for Computational Linguistics: Human Language Technologies*, pages 547–564, Stroudsburg, PA, USA. Association for Computational Linguistics.

Majid Asgari-Bidhendi, Ali Hadian, and Behrouz Minaei-Bidgoli. 2019. FarsBase: The Persian knowledge graph. *Semantic Web*, 10(6):1169–1196.

Adel Atef, Bassam Mattar, Sandra Sherif, Eman Elrefai, and Marwan Torki. 2020. AQAD: 17,000+ Arabic Questions for Machine Comprehension of Text. *Proceedings of IEEE/ACS International Conference on Computer Systems and Applications, AICCSA*, 2020-Novem:1–6.

B. Barla Cambazoglu, Mark Sanderson, Falk Scholer, and Bruce Croft. 2020. A Review of Public Datasets in Question Answering Research. *ACM SIGIR Forum*, 54(2):1–23.

Jonathan H Clark, Eunsol Choi, Michael Collins, Dan Garrette, Tom Kwiatkowski, Vitaly Nikolaev, and Jennimaria Palomaki. 2020. TyDi QA: A Benchmark for Information-Seeking Question Answering in Typologically Diverse Languages. *Transactions of the Association for Computational Linguistics*, 8:454–470, December.

Peter Clark, Isaac Cowhey, Oren Etzioni, Tushar Khot, Ashish Sabharwal, Carissa Schoenick, and Oyvind

Tafjord. 2018. Think you have Solved Question Answering? Try ARC, the AI2 Reasoning Challenge.

Irene Cramer, Jochen L. Leidner, and Dietrich Klakow. 2006. Building an evaluation corpus for German question answering by harvesting wikipedia. *Proceedings of the 5th International Conference on Language Resources and Evaluation, LREC 2006*:1514–1519.

Danilo Croce, Alexandra Zelenanska, and Roberto Basili. 2018. Neural Learning for Question Answering in Italian. *AI*IA 2018 – Advances in Artificial Intelligence*, 11298:389–402.

Yiming Cui, Ting Liu, Wanxiang Che, Li Xiao, Zhipeng Chen, Wentao Ma, Shijin Wang, and Guoping Hu. 2020. A span-extraction dataset for Chinese machine reading comprehension. *EMNLP-IJCNLP 2019 - 2019 Conference on Empirical Methods in Natural Language Processing and 9th International Joint Conference on Natural Language Processing, Proceedings of the Conference*:5883–5889.

Yiming Cui, Ting Liu, Zhipeng Chen, Wentao Ma, Shijin Wang, and Guoping Hu. 2019. Dataset for the first evaluation on Chinese machine reading comprehension. *LREC 2018 - 11th International Conference on Language Resources and Evaluation*:2721–2725.

Yiming Cui, Ting Liu, Zhipeng Chen, Shijin Wang, and Guoping Hu. 2016. Consensus attention-based neural networks for Chinese reading comprehension. *COLING 2016 - 26th International Conference on Computational Linguistics, Proceedings of COLING 2016: Technical Papers*:1777–1786.

Martin d'Hoffschmidt, Wacim Belblidia, Quentin Heinrich, Tom Brendlé, and Maxime Vidal. 2020. FQuAD: French Question Answering Dataset. In *Findings of the Association for Computational Linguistics: EMNLP 2020*, pages 1193–1208, Stroudsburg, PA, USA. Association for Computational Linguistics.

Daria Dzendzik, Carl Vogel, and Jennifer Foster. 2021. English Machine Reading Comprehension Datasets: A Survey. In *arXiv:2101.10421v2 [cs.CL]*.

Pavel Efimov, A v Chertok, Leonid Boytsov, and Pavel Braslavski. 2020. SberQuAD – Russian Reading Comprehension Dataset : Description and Analysis. In *International Conference of the Cross-Language Evaluation Forum for European Languages*, pages 3–15.

Romina Etezadi and Mehrnoush Shamsfard. 2020. PeCoQ: A Dataset for Persian Complex Question Answering over Knowledge Graph. In *2020 11th International Conference on Information and Knowledge Technology (IKT)*, pages 102–106. IEEE, December.

Dirk Groeneveld, Tushar Khot, Mausam, and Ashish Sabharwal. 2020. A Simple Yet Strong Pipeline for HotpotQA. In *Proceedings of the 2020 Conference on Empirical Methods in Natural Language Processing (EMNLP)*, pages 8839–8845, Stroudsburg, PA, USA. Association for Computational Linguistics.

Aditya Gupta, Jiacheng Xu, Shyam Upadhyay, Diyi Yang, and Manaal Faruqui. 2021. Disfl-QA: A Benchmark Dataset for Understanding Disfluencies in Question Answering. In *Findings of the Association for Computational Linguistics: ACL-IJCNLP 2021*, pages 3309–3319, Stroudsburg, PA, USA. Association for Computational Linguistics.

Deepak Gupta, Surabhi Kumari, Asif Ekbal, and Pushpak Bhattacharyya. 2018. MMQA : A Multi-domain Multi-lingual Question-Answering Framework for English and Hindi. In *Proceedings of the Eleventh International Conference on Language Resources and Evaluation (LREC 2018)*, pages 2777–2784.

Momchil Hardalov, Ivan Koychev, and Preslav Nakov. 2019. Beyond English-only reading comprehension: Experiments in zero-shot multilingual transfer for Bulgarian. *International Conference Recent Advances in Natural Language Processing, RANLP*, 2019-Septe:447–459.

Wei He, Kai Liu, Jing Liu, Yajuan Lyu, Shiqi Zhao, Xinyan Xiao, Yuan Liu, Yizhong Wang, Hua Wu, Qiaoqiao She, Xuan Liu, Tian Wu, and Haifeng Wang. 2018. DuReader: a Chinese Machine Reading Comprehension Dataset from Real-world Applications. In *Proceedings of the Workshop on Machine Reading for Question Answering*, pages 37–46, Stroudsburg, PA, USA. Association for Computational Linguistics.

Michael A. Hedderich, Lukas Lange, Heike Adel, Jannik Strötgen, and Dietrich Klakow. 2021. A Survey on Recent Approaches for Natural Language Processing in Low-Resource Scenarios. In *Proceedings of the 2021 Conference of the North American Chapter of the Association for Computational Linguistics: Human Language Technologies*, pages 2545–2568, Stroudsburg, PA, USA. Association for Computational Linguistics.

Quentin Heinrich, Gautier Viaud, Wacim Belblidia, and Illuin Technology. 2021. FQuAD2.0: French Question Answering and knowing that you know nothing. In *arXiv:2109.13209 [cs.CL]*.

Felix Hill, Antoine Bordes, Sumit Chopra, and Jason Weston. 2016. The Goldilocks principle: Reading children's books with explicit memory representations. *4th International Conference on*

*Learning Representations, ICLR 2016 - Conference Track Proceedings*:1–13.

Zhen Huang, Shiyi Xu, Minghao Hu, Xinyi Wang, Jinyan Qiu, Yongquan Fu, Yuncai Zhao, Yuxing Peng, and Changjian Wang. 2020. Recent Trends in Deep Learning Based Open-Domain Textual Question Answering Systems. *IEEE Access*, 8:94341–94356.

Walaa Saber Ismail and Masun Nabhan Homsi. 2018. DAWQAS: A Dataset for Arabic Why Question Answering System. *Procedia Computer Science*, 142:123–131.

Yimin Jing, Deyi Xiong, and Yan Zhen. 2020. BiPaR: A bilingual parallel dataset for multilingual and cross-lingual reading comprehension on novels. *EMNLP-IJCNLP 2019 - 2019 Conference on Empirical Methods in Natural Language Processing and 9th International Joint Conference on Natural Language Processing, Proceedings of the Conference*:2452–2462.

Rachel Keraron, Guillaume Lancrenon, Mathilde Bras, Gilles Moyse, Thomas Scialom, and Jacopo Staiano. 2020. Project PIAF : Building a Native French Question-Answering Dataset. In *Proceedings of the 12th Language Resources and Evaluation Conference*, number May, pages 5481–5490.

Omri Keren and Omer Levy. 2021. ParaShoot: A Hebrew Question Answering Dataset. In *Proceedings of the 3rd Workshop on Machine Reading for Question Answering*, pages 106–112.

Daniel Khashabi, Arman Cohan, Siamak Shakeri, Pedram Hosseini, Pouya Pezeshkpour, Malihe Alikhani, Moin Aminnaseri, Marzieh Bitaab, Faeze Brahman, Sarik Ghazarian, Mozhdeh Gheini, Arman Kabiri, Rabeeh Karimi Mahabagdi, Omid Memarrast, Ahmadreza Mosallanezhad, Erfan Noury, Shahab Raji, Mohammad Sadegh Rasooli, Sepideh Sadeghi, et al. 2021. ParsiNLU: A Suite of Language Understanding Challenges for Persian. *Transactions of the Association for Computational Linguistics*, 9:1147–1162, October.

Vladislav Korablinov and Pavel Braslavski. 2020. RuBQ : A Russian Dataset for Question Answering over Wikidata. In *arXiv:2005.10659*.

Tom Kwiatkowski, Jennimaria Palomaki, Olivia Redfield, Michael Collins, Ankur Parikh, Chris Alberti, Danielle Epstein, Illia Polosukhin, Jacob Devlin, Kenton Lee, Kristina Toutanova, Llion Jones, Matthew Kelcey, Ming-Wei Chang, Andrew M. Dai, Jakob Uszkoreit, Quoc Le, and Slav Petrov. 2019. Natural Questions: A Benchmark for Question Answering Research. *Transactions of the Association for Computational Linguistics*, 7:453–466.

Kyungjae Lee, Kyoungho Yoon, Sunghyun Park, and Seung Won Hwang. 2019. Semi-supervised training data generation for multilingual question answering. *LREC 2018 - 11th International Conference on Language Resources and Evaluation*:2758–2762.

Patrick Lewis, Barlas Oguz, Ruty Rinott, Sebastian Riedel, and Holger Schwenk. 2020. MLQA: Evaluating Cross-lingual Extractive Question Answering. In *Proceedings of the 58th Annual Meeting of the Association for Computational Linguistics*, pages 7315–7330, Stroudsburg, PA, USA. Association for Computational Linguistics.

Seungyoung Lim, Myungji Kim, and Jooyoul Lee. 2019. KorQuAD1.0: Korean QA Dataset for Machine Reading Comprehension. In *arXiv:1909.07005v2 [cs.CL]*, pages 0–4.

Jiahua Liu, Yankai Lin, Zhiyuan Liu, and Maosong Sun. 2019a. XQA: A Cross-lingual Open-domain Question Answering Dataset. In *Proceedings of the 57th Annual Meeting of the Association for Computational Linguistics*, pages 2358–2368, Stroudsburg, PA, USA. Association for Computational Linguistics.

Qianying Liu, Sicong Jiang, Yizhong Wang, and Sujian Li. 2020. LiveQA: A Question Answering Dataset Over Sports Live. *Lecture Notes in Computer Science (including subseries Lecture Notes in Artificial Intelligence and Lecture Notes in Bioinformatics)*, 12522 LNAI(c):316–328.

Shanshan Liu, Xin Zhang, Sheng Zhang, Hui Wang, and Weiming Zhang. 2019b. Neural machine reading comprehension: Methods and trends. *Applied Sciences (Switzerland)*, 9(18).

Shayne Longpre, Yi Lu, and Joachim Daiber. 2020. MKQA: A Linguistically Diverse Benchmark for Multilingual Open Domain Question Answering. In *arXiv:2007.15207v2 [cs.CL]*.

Hussein Mozannar, Elie Maamary, Karl el Hajal, and Hazem Hajj. 2019. Neural Arabic Question Answering. In *Proceedings of the Fourth Arabic Natural Language Processing Workshop*, number 1, pages 108–118, Stroudsburg, PA, USA. Association for Computational Linguistics.

Kiet Nguyen, Vu Nguyen, Anh Nguyen, and Ngan Nguyen. 2020a. A Vietnamese Dataset for Evaluating Machine Reading Comprehension. In *Proceedings of the 28th International Conference on Computational Linguistics*, pages 2595–2605, Stroudsburg, PA, USA. International Committee on Computational Linguistics.

Kiet van Nguyen, Khiem Vinh Tran, Son T. Luu, Anh Gia-Tuan Nguyen, and Ngan Luu-Thuy Nguyen. 2020b. Enhancing Lexical-Based Approach With External Knowledge for Vietnamese Multiple-Choice Machine Reading Comprehension. *IEEE Access*, 8:201404–201417.

Kyosuke Nishida, Itsumi Saito, Atsushi Otsuka, Hisako Asano, and Junji Tomita. 2018. Retrieve-and-read: Multi-task learning of information retrieval and reading comprehension. *International Conference on Information and Knowledge Management, Proceedings*:647–656.

Pranav Rajpurkar, Jian Zhang, Konstantin Lopyrev, and Percy Liang. 2016. SQuAD: 100,000+ Questions for Machine Comprehension of Text. In *Proceedings of the 2016 Conference on Empirical Methods in Natural Language Processing*, pages 2383–2392, Stroudsburg, PA, USA. Association for Computational Linguistics.

Julian Risch and Malte Pietsch. 2017. GermanQuAD and GermanDPR: Improving Non-English Question Answering and Passage Retrieval. In *arXiv:2104.12741v1 [cs.CL]*.

Anna Rogers, Matt Gardner, and Isabelle Augenstein. 2021. QA Dataset Explosion: A Taxonomy of NLP Resources for Question Answering and Reading Comprehension. In *arXiv:2107.12708 [cs.CL]*, pages 1–38.

Ivan Rybin, Vladislav Korablinov, and Pavel Efimov. 2021. RuBQ 2.0 : An Innovated Russian Question Answering Dataset. In *Eighteenth Extended Semantic Web Conference - Resources Track*, pages 1–14.

Helio Fonseca Sayama, Anderson Vicoso Araujo, and Eraldo Rezende Fernandes. 2019. FaQuAD: Reading comprehension dataset in the domain of brazilian higher education. *Proceedings - 2019 Brazilian Conference on Intelligent Systems, BRACIS 2019*(March):443–448.

Chih Chieh Shao, Trois Liu, Yuting Lai, Yiying Tseng, and Sam Tsai. 2018. DRCD: a Chinese Machine Reading Comprehension Dataset. In *arXiv:1806.00920v3 [cs.CL]*.

Wissam Siblini, Charlotte Pasqual, Axel Lavielle, Mohamed Challal, and Cyril Cauchois. 2021. Multilingual Question Answering applied to Conversational Agents. In *arXiv:1910.04659v2 [cs.CL]*.

Yasunobu Sumikawa, Masaaki Fujiyoshi, Hisashi Hatakeyama, and Masahiro Nagai. 2019. An FAQ dataset for E-learning system used on a Japanese University. *Data in Brief*, 25:104001.

Norio Takahashi, Tomohide Shibata, Daisuke Kawahara, and Sadao Kurohashi. 2019. Machine Comprehension Improves Domain-Specific Japanese Predicate-Argument Structure Analysis. In *Proceedings of the 2nd Workshop on Machine Reading for Question Answering*, pages 98–104, Stroudsburg, PA, USA. Association for Computational Linguistics.

Kiet van Nguyen, Tin van Huynh, Duc-Vu Nguyen, Anh Gia-Tuan Nguyen, and Ngan Luu-Thuy Nguyen. 2020. New Vietnamese Corpus for Machine Reading Comprehension of Health News Articles. In *arXiv:2006.11138v2 [cs.CL]*, volume 1, pages 1–41.

Hadi Veisi and Hamed Fakour Shandi. 2020. A Persian Medical Question Answering System. *International Journal on Artificial Intelligence Tools*, 29(06):2050019, September.

Qizhe Xie, Guokun Lai, Zihang Dai, and Eduard Hovy. 2018. Large-scale Cloze Test Dataset Created by Teachers. In *Proceedings of the 2018 Conference on Empirical Methods in Natural Language Processing*, pages 2344–2356, Stroudsburg, PA, USA. Association for Computational Linguistics.

Changchang Zeng, Shaobo Li, Qin Li, Jie Hu, and Jianjun Hu. 2020a. A Survey on Machine Reading Comprehension—Tasks, Evaluation Metrics and Benchmark Datasets. *Applied Sciences*, 10(21):7640, October.

Changchang Zeng, Shaobo Li, Qin Li, Jie Hu, and Jianjun Hu. 2020b. A survey on machine reading comprehension—tasks, evaluation metrics and benchmark datasets. *Applied Sciences (Switzerland)*, 10(21):1–57.

Qiyuan Zhang, Lei Wang, Sicheng Yu, Shuohang Wang, Yang Wang, Jing Jiang, and Ee-Peng Lim. 2021a. NOAHQA: Numerical Reasoning with Interpretable Graph Question Answering Dataset. In *Findings of the Association for Computational Linguistics: EMNLP 2021*, pages 4147–4161.

Taolin Zhang, Chengyu Wang, Minghui Qiu, Bite Yang, Zerui Cai, Xiaofeng He, and Jun Huang. 2021b. Knowledge-Empowered Representation Learning for Chinese Medical Reading Comprehension: Task, Model and Resources. In *Findings of the Association for Computational Linguistics: ACL-IJCNLP 2021*, pages 2237–2249, Stroudsburg, PA, USA. Association for Computational Linguistics.